\documentclass[twocolumn]{article}
\usepackage[letterpaper]{geometry}
\usepackage{spconf}
\usepackage{amsmath,epsfig}
\usepackage[noadjust]{cite}

\usepackage{hyperref}
\hypersetup{
    colorlinks=true,
    linkcolor=blue,
    filecolor=magenta,      
    urlcolor=cyan,
}
\usepackage{fancyhdr}
\usepackage[skip=0pt]{caption}
\usepackage{graphicx}
\DeclareGraphicsExtensions{.pdf,.png,.jpg}

\usepackage{euler}
\usepackage{amsfonts}

\usepackage{algorithm}
\usepackage[noend]{algpseudocode}
\makeatletter
\def\BState{\State\hskip-\ALG@thistlm}
\makeatother

\usepackage{float}
\newfloat{algorithm}{t}{lop}

\newcommand{\Jo}{\mathfrak{\textbf{J}}}
\newcommand{\Sf}{\mathfrak{\textbf{S}}}
\newcommand{\Df}{\mathfrak{\textbf{D}}}
\newcommand{\Vs}{\theta}
\newcommand{\Xy}{\mathfrak{\textbf{x}}}
\newcommand{\Dz}{\omega}

\begin{document}\sloppy

\def\x{{\mathbf x}}
\def\L{{\cal L}}

\title{ Variational Disparity Estimation Framework for Plenoptic Images }
%
\name{Trung-Hieu Tran, Zhe Wang, Sven Simon}
\address{  Department of Parallel Systems\\
University of Stuttgart, Stuttgart, Germany \\
trung.hieu.tran@ipvs.uni-stuttgart.de}

\maketitle

\begin{abstract}
This paper presents a computational framework for accurately estimating the 
disparity map of plenoptic images. The proposed framework is based on 
the variational principle and provides intrinsic sub-pixel precision. 
The light-field motion tensor introduced in the framework allows us to combine 
advanced robust data terms as well as provides explicit 
treatments for different color channels. A warping strategy is embedded in our 
framework for tackling the large displacement problem. We also show that by 
applying a simple regularization  term and a guided median 
filtering, the accuracy of displacement field at occluded area could be greatly 
enhanced. We demonstrate the excellent performance of the proposed framework 
by intensive comparisons with the Lytro software and contemporary approaches on 
both synthetic and real-world datasets.

\end{abstract}
\begin{keywords}
light-field, correspondence, disparity, plenoptic, variational framework
\end{keywords}
\section{introduction}
\label{sec:intro}

The last decade saw the dramatically increasing attention of the research 
community to light-field photography. Light-field acquisition provides a richer 
content capturing method compared to traditional photography for not only 
acquiring the spatial but also the directional information of the scene. Various 
approaches were provided for capturing light-field such as multi-camera 
array~\cite{Wilburn2005}, programmatically  moving camera (also known as gantry 
setup), and microlens-array camera~\cite{Adelson1992}. Among them the 
lenslet-based or plenoptic camera~\cite{Adelson1992,Ng2005,Lumsdaine2009} 
provides the most convenient and efficient way to acquire light-field. On the 
one hand, it is far less hardware intensive compared to multi-camera array. On 
the other hand, unlike gantry which captures only a still scene, it could 
capture both still and dynamic sceneries. 

In order to take the most benefits from captured light-field, the 
disparity map estimation is the most important task that must be considered. 
Accurate estimation of disparity map will enable the better visualization and 
manipulation of light-field data. However, it still remains as a challenging 
problem 
for plenoptic images. Compared with conventional stereo images, plenoptic 
sub-aperture images possess a very narrow-baseline configuration with typical 
disparity range between -1 and 1 pixel. For computing an accurate result, 
discrete label-based approaches~\cite{Tao2015,Wang2015,Jaesik2015} 
require the previous knowledge of sub-pixel 
disparity unit for initial disparity map estimation and then requires 
exhaustive adjustment of regularization and refinement parameters in order to 
acquire 
a smooth and precise solution. This fact, therefore, limits the use of these 
approaches for practical applications.

\begin{figure}
\centering
\begin{minipage}[b]{.95\linewidth}
\begin{minipage}[b]{.32\linewidth}
\begin{minipage}[b]{\linewidth}
  \centering
  \includegraphics[width=\textwidth]{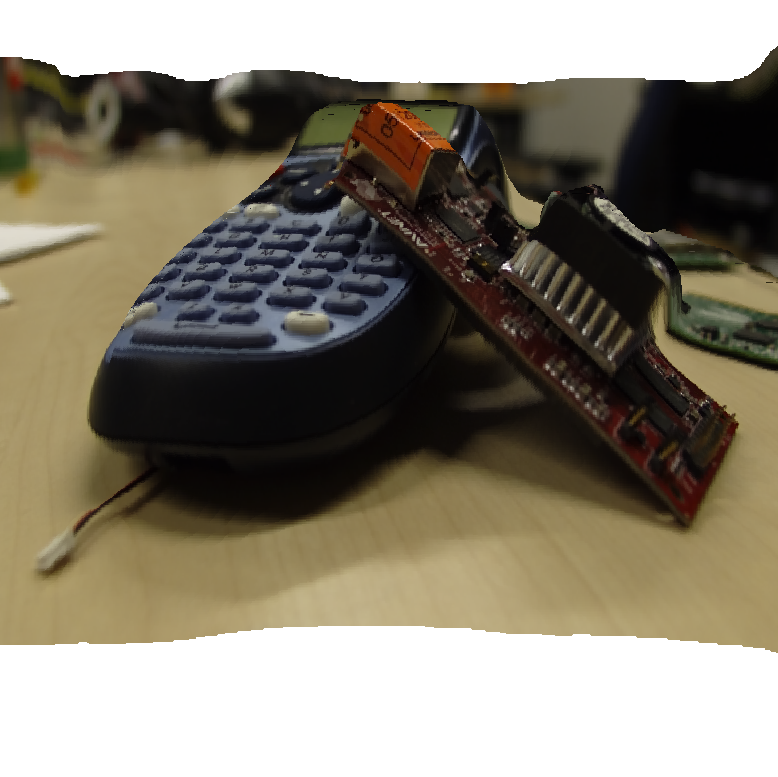}
\end{minipage}
\begin{minipage}[b]{\linewidth}
  \centering
  \includegraphics[width=\textwidth]{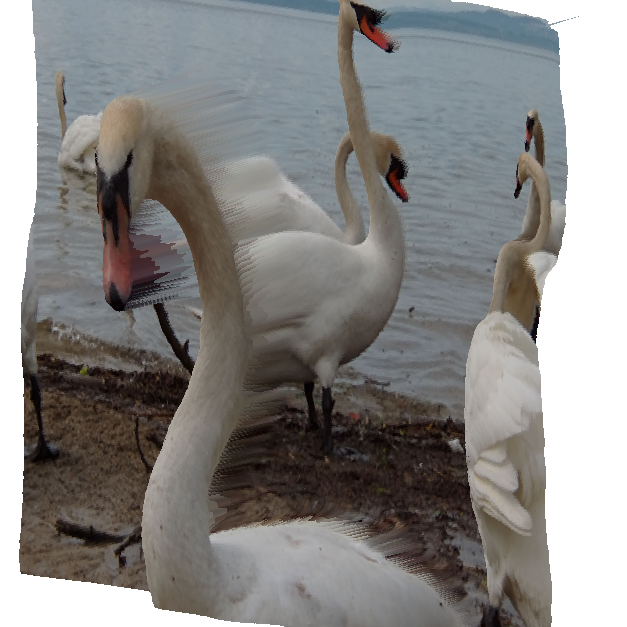}
\end{minipage}
  \centerline{3D view}\medskip
\end{minipage}
\hfill
\begin{minipage}[b]{.315\linewidth}
\begin{minipage}[b]{\linewidth}
  \centering
  \includegraphics[width=\textwidth]{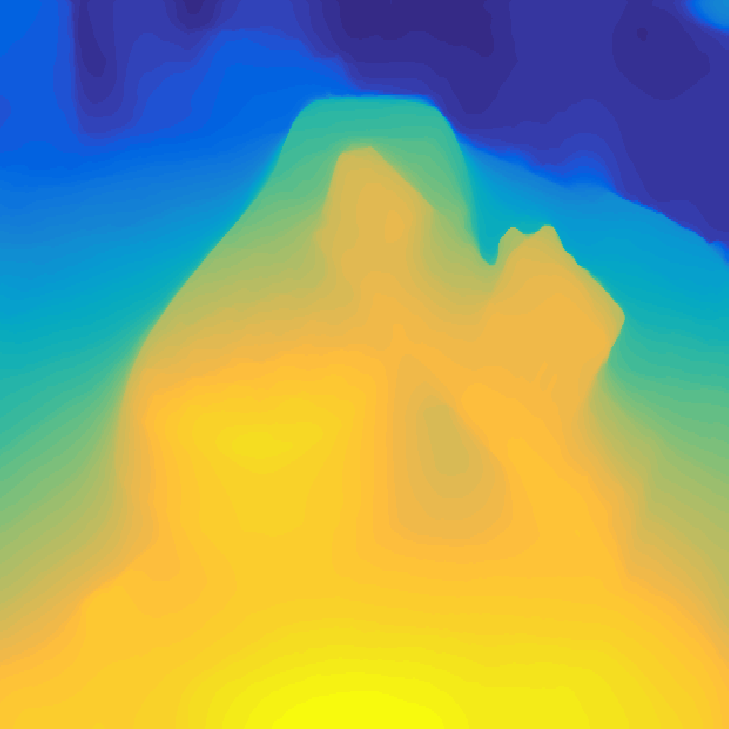}
\end{minipage}
\begin{minipage}[b]{\linewidth}
  \centering
  \includegraphics[width=\textwidth]{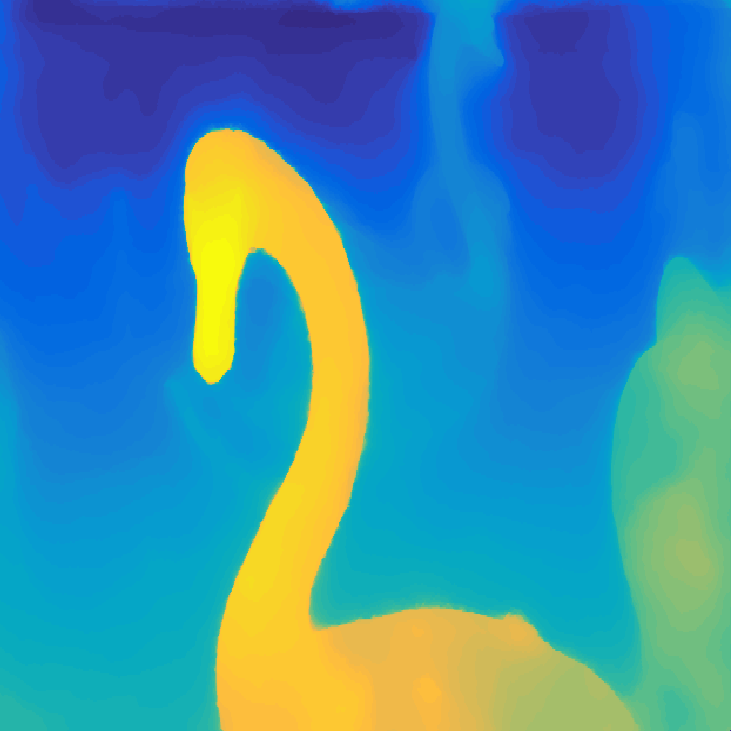}
\end{minipage}
  \centerline{Disparity}\medskip
\end{minipage}
\hfill
\begin{minipage}[b]{.32\linewidth}
\begin{minipage}[b]{\linewidth}
  \centering
  \includegraphics[width=\textwidth]{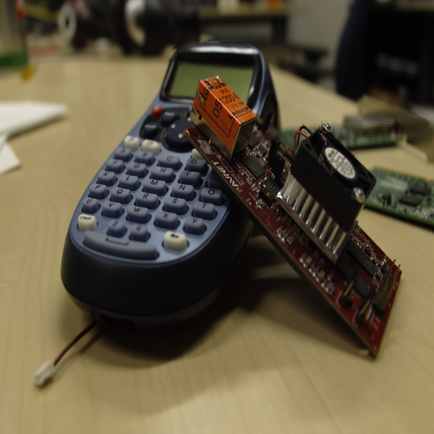}
\end{minipage}
\begin{minipage}[b]{\linewidth}
  \centering
  \includegraphics[width=\textwidth]{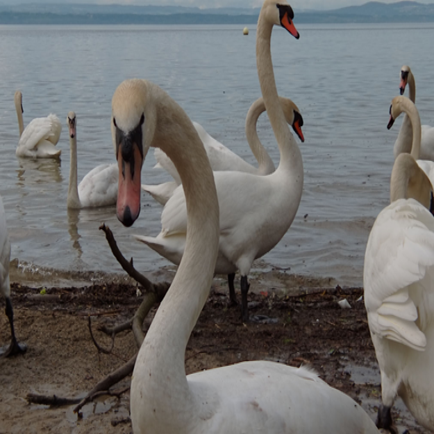}
\end{minipage}
  \centerline{Central view}\medskip
\end{minipage}
\end{minipage}
\caption{Real world result. \textit{top} scene from our laboratory captured 
light-field. \textit{bottom} scene from EPFL dataset~\cite{epfl2016}.}
\label{fig:real_common}
\end{figure}

In this paper, we present a computational framework for robust light-field 
disparity estimation. We formulate the problem using a continuous variational 
framework that allows an intrinsic sub-pixel precision and introduce a joint 
light-field motion tensor for combining different advanced data terms. A 
coarse-to-fine warping strategy is embedded in our framework for 
dealing with the problem of large displacement in sub-aperture images at 
directional borders. In addition, we also show that by applying a guided median 
filter appropriately, the accuracy of displacement field at occluded areas 
could be greatly enhanced. Fig.~\ref{fig:real_common} demonstrates the 
results of our proposed framework on Lytro Illium light-field data.

In the following, we provide a brief introduction on the mathematical 
representation of light-field and literature review on 
related works. 
Our light-field variational framework is then 
formulated and presented in Section~\ref{sec:modeling}. The post-processing 
technique for sharpening 
disparity discontinues will be discussed in Section~\ref{sec:post_processing}. 
In Section~\ref{sec:experiment}, we provide the results of our intensive 
experiments on both synthetic dataset and real world dataset. Both quantitative 
and qualitative comparisons with state-of-the art approaches are also provided 
in this section.

\subsection{Light-field}
\label{subsec:lightfield}
Light-field is a 4D parameterization of the plenoptic function, it could be 
visually described as a ray indexed by its intersection with two parallel 
planes. 
\begin{equation}
L : \Omega \times \Pi \rightarrow \mathbb{R},  (\Xy,\Vs) \rightarrow L(\Xy,\Vs)
\end{equation}
where $\Xy =(x,y)^T$ and $\Vs = (s,t)^T$ denote coordinate pairs in the image 
plane $\Omega \subset \mathbb{R}^2$ and in the lens plane $\Pi \subset 
\mathbb{R}^2$. For lenslet-based light-field camera, $\Omega$ and $\Pi$ 
correspond to the spatial and directional coordinates respectively.

Given a light-field $L$, spatial information could be obtained from one 
direction by keeping the directional component $\Vs$ unchanged and varying over 
all spatial domain $\Omega$. Such spatial information gives us a 
sub-aperture image (or a view) of the captured scene. The number of 
sub-aperture images depends on the configuration of lenslet-based camera with 
a consideration of the trade-off between spatial and angular 
resolution~\cite{Lumsdaine2009}. For example, contemporary cameras such as 
Lytro Illium provide up to $15 \times 15$ sub-aperture images with the same 
amount of 
reduction in spatial domain. These dense multiple overlapping sub-aperture 
images show a clear relation to multi-stereo problems or could be referred as 
a correspondence matching problem. This point of view motivates us to propose 
this computational framework. 

\subsection{Related works}
\label{subsec:related}
Bishop and Favaro~\cite{Bishop2012} constructed an image formation for 
lenslet-based light-field camera and embedded it in their iterative method for 
multi-view stereo problems. Wanner et al.~\cite{Wanner2012,Wanner2014a} 
computed a depth map through estimating the vertical and horizontal slopes in 
epipolar planes. They then proposed a global optimization scheme using 
a variational framework for the final depth map. Yu et al.~\cite{Yu2013} 
explored geometric structures of 3D 
lines in ray space and computed disparity maps through line matching between 
the sub-aperture images. Tao et al.~\cite{Tao2015} used both correspondence and 
defocus cue to estimate disparity maps with intention to complement the 
disadvantages of each other. Wang et al.~\cite{Wang2015} proposed a depth 
estimation algorithm that treats occlusion explicitly. Jeon et 
al.~\cite{Jaesik2015} proposed the method based on the phase shift theorem to 
deal with narrow baseline multi-view images. 
In~\cite{Tao2015,Wang2015,Jaesik2015}, the proposed approaches 
came up with a discrete label depth map, and a multi-label optimization 
method was applied in order to regularize and refine the estimated disparity 
map. 

Compared with these label-based approaches, continuous modeling 
approaches~\cite{Heber2013,Heber2014} have advantages because it could 
provide an intrinsic sub-pixel precision and do not require previous knowledge 
on sub-pixel displacement units. In ~\cite{Heber2013}, Heber and Pocks 
proposed a method that is based on active wave-font sampling and the 
variational principle. In ~\cite{Heber2014}, a method is proposed to estimate 
disparity maps using the low-rank structure regularization to align the 
sub-aperture images.

Compared with previous works, especially variational 
approaches~\cite{Wanner2014a,Heber2013},
our contribution is two-fold. Firstly, we introduce a light-field motion 
tensor that allows us to take advantages of different constancy assumptions
and explicitly define the contribution of different color channels. 
Secondly, we proposed to enhance the discontinues of disparity 
field at occluded areas by appropriately applying guided median filtering. 
Experimental results show the great accuracy achievement when compared with 
previous approaches.

\section{proposed framework}
\label{sec:modeling}
In this section, we formulate the proposed light-field disparity 
estimation framework, which is based on a variational model. We begin with 
the general form of a variational problem which is usually written as follows.
\begin{equation}
\operatornamewithlimits{argmin}_{\Dz} E(\Dz) = \int\limits_{\Omega} \Df(L,\Dz) 
+ 
\alpha\Sf(L,\Dz)d\Xy
\end{equation}
where $\Df$ refers to a data term, $\Sf$ refers to a smoothness term or 
a regularization term and $\alpha>0$ denotes a smoothness weight. The data 
function $\Df$ penalizes the deviations of estimated solution $\Dz$ from the 
true solution with respect to the constraints on the data $L$. Such 
constraints are known as the constancy assumption in optical flow 
literature~\cite{Bruhn2006},\cite{Zimmer2011}. 
The regularization function $\Sf$ takes into account the neighborhood 
information and guarantees the smoothness of the solution.

We introduce unit displacement $\Dz: (x,y) \rightarrow \mathbb{R}$ which 
denotes the disparity between the pixels in the central view and in its direct 
east neighbor view. Based on $\Dz$ we could define the constancy 
assumption across the light-field $L(\Xy,\Vs)$.

\begin{equation}
L(\Xy,\Vs_0) = L(\Xy + \kappa\Vs_i\Dz,\Vs_i)
\label{eq:gray_constant}
\end{equation}
Here, $\Vs_0$ denotes the central view index and $\Vs_i$ denotes an arbitrary 
view index within the directional domain $\Pi$. A parameter $\kappa = 
\kappa_{\Omega} \kappa_{\Pi} = \begin{bmatrix}k &0\\0 & 1\end{bmatrix}$, $k \in 
\mathbb{R}$, 
compensates the relative difference 
between the horizontal and 
vertical components of directional ($\kappa_{\Pi}$) and spatial 
($\kappa_{\Omega}$) coordinates. Without loss of generality and to simplify the 
exposition, we will absorb $\kappa$ into $\theta$. Following sections will 
discuss in detail the various settings of the data term and the smoothness term.

\subsection{Data term}
\textit{Intensity constancy assumption} The Eq.~\ref{eq:gray_constant} 
actually presents the constancy assumption of pixel intensity that assumes 
the intensity of corresponding pixels between two sub-aperture images is 
the same. Supposes $\Dz$ is small, the first-order Taylor expansion gives us 
the following
\begin{equation}
0 = \Dz\Vs_i^T \nabla_2L + |\Vs_i|L_\Vs = w^T\nabla_{\Vs_i} L
\label{eq:taylor}
\end{equation}
where $\nabla_2 = (\sigma_x,\sigma_y)^T$ denotes the spatial gradient operator, 
$L_\Vs$ denotes the directional derivative with direction $\Vs_i$, $w=(\Dz,1)^T$ 
and $\nabla_{\Vs_i} L = (\Vs_i^T \nabla_2L,|\Vs_i|L_\Vs)^T$

Using Eq.~\ref{eq:taylor}, we could derive a data term that takes into account 
the intensity constancy assumption as follows.
\begin{equation}
\begin{split}
\Df_g(L,\Dz) =&  \operatornamewithlimits{\sum_{c=1}}^3 \Psi_g \Big( 
\sum\limits_{\Vs_i\in\Pi} \big( L(\Xy + \Vs_i \Dz,\Vs_i) - L(\Xy,\Vs_0) \big)^2 
\Big)\\
 		 \approx& \operatornamewithlimits{\sum_{c=1}}^3 \Psi_g \Big( 
\sum\limits_{\Vs_i\in\Pi} w^T\Jo_{g,i}^c w \Big)
 		 = \operatornamewithlimits{\sum_{c=1}}^3 \Psi_g \big( 
w^T\Jo_{g}^c w \big)
\end{split}
\label{eq:gray_data}
\end{equation}
where $\Jo_{g,i}^c = \nabla_{\Vs_i} L^c \nabla_{\Vs_i}^T L^c $ denotes a 
light-field 
motion tensor for a single color channel, and $L^c$ denotes one color channel 
of the captured light-field $L$. We define a joint light-field 
motion tensor 
$\Jo_g^c$ for all views $\Vs_i \in \Pi$ as in Eq.~\ref{eq:gray_data}. 
$\Psi_g(s)$ 
is a positive defined robustification function that helps in reducing the 
outliers. Here we 
choose $\Psi_g(s) = \sqrt{s+\epsilon_g}$, with $\epsilon_g > 0$ serves as 
a small 
regularization parameter, which also allows the derivative of $\Psi_g$ 
available 
when $s=0$. This $L_1$ norm is known to be better in handling outliers 
caused by noise and occlusions.

\textit{Gradient constancy assumption} In addition to intensity, it also makes 
sense to assume that the gradient of corresponding pixels is also unchanged. 
In the same manner, the gradient constancy assumption data term could be 
described as follows.
\begin{equation}
\begin{split}
\Df_G(L,\Dz) =& \operatornamewithlimits{\sum_{c=1}}^3 \Psi_G\Big( 
\sum\limits_{\Vs_i\in\Pi} (w^T\Jo_{G_x,i}^cw + w^T\Jo_{G_y,i}^cw ) \Big)\\
		   =& \operatornamewithlimits{\sum_{c=1}}^3 \Psi_G\Big( w^T\Jo_G^cw \Big)
\end{split}
\end{equation}
where $\Jo_{G_{*},i}^c = \nabla_\Vs L_{*}^c \nabla_\Vs^T L_{*}^c$ and $L_{*}^c$ 
define the derivative of the light-field $L$ on color chanel $c$.

\textit{Color space} Both the \textit{Red Green Blue} (RGB) and the \textit{Hue 
Saturation Value} (HSV) color spaces contain different characteristics that 
could be exploited for providing a constancy assumption. Both of them are 
experimented in our framework. Notes that, 
we apply the join robustification in the case of RGB because of the mutual 
relationship between these three channels. For HSV, the separate $L_1$ norms are 
used instead, since each color channel in this case contain information that is 
not encoded in other channels~\cite{Zimmer2011}. 

Combining both constancy assumptions, we have the final form of the data 
function. 
For the RGB color space:
\begin{equation}
\begin{split}
\Df(L,\Dz) =  \Psi_g\Big( \operatornamewithlimits{\sum_{c=1}}^3 w^T\Jo_g^cw 
\Big)
         + \gamma\ \Psi_G \Big( \operatornamewithlimits{\sum_{c=1}}^3  w^T\Jo_G^cw  \Big)
\end{split}
\end{equation}
And in the case of the HSV color space.
\begin{equation}
\Df(L,\Dz) = \operatornamewithlimits{\sum_{c=1}}^3 \Psi_g\big( w^T\Jo_g^cw 
\big) + \gamma\ \operatornamewithlimits{\sum_{c=1}}^3 \Psi_G\big( w^T\Jo_G^cw 
\big)
\end{equation}
Parameter $\gamma$ allows us to adjust the importance of gradient constancy 
data term. In following sections, we focus on HSV data term and the RGB data 
term 
could be simply derived with a similar manner.

\subsection{Smoothness term}
The light-field motion tensor possesses a strong energy in crowd texture 
area. However, it is weak in homogeneous area and could lead to wrong 
estimations. 
The smoothness term in this case plays a crucial role because it spreads the 
information through the neighborhood and provides a smooth disparity field. 
There are various regularization functions that could fit into our framework. 
The 
most well known one is $L_2$ Total Variation ($TV-L_2$) that penalizes the 
variation of $\Dz$ in two spatial directions using a quadratic function.
\begin{equation}
S(\Dz) = |\nabla_2 \Dz|^2
\end{equation}

Despite providing a smooth solution, this regularizer also blurs the disparity 
edges and reduces the accuracy of the solution. One solution to it is an 
isotropic image-driven smoothness term defined with a weighting function 
$g(\Xy)$
\begin{equation}
S(\Dz) = g(\Xy)|\nabla_2 \Dz|^2
\end{equation}
Here, $g(\Xy)=\frac{1}{\sqrt{|\nabla_2 L(\Xy,\Vs_0)|^2 +\epsilon}}$ allows 
reducing smoothness at image edge therefore results in a 
sharper disparity edge. One problem with this regularization term is the 
over-segmentation especially in dense texture area. For that reason, we choose 
the non-quadratic penalizer also known as $TV-L_1$ that allows piece-wise 
smoothness in disparity field.
\begin{equation}
\Sf(\Dz) = \Psi_s( |\nabla_2\Dz|^2 )
\end{equation}
where $\Psi_s(s) = \sqrt{s+\epsilon_s}$, $\epsilon_s>0$. It also notes that 
more sophisticated regulizers such as joint image- and 
flow-driven~\cite{Zimmer2011} or an-isotropic diffusion 
tensor~\cite{Heber2013,Ranftl2012} are also possible in our framework.
\subsection{Optimization}
Combining both data and smoothness terms, we have a final global variational 
energy function.
\begin{equation}
\begin{split}
E(\Dz) = \int\limits_{\Omega} & \operatornamewithlimits{\sum_{c=1}}^3 \Psi_g\big( w^T\Jo_g^cw \big) + \gamma\ \operatornamewithlimits{\sum_{c=1}}^3 \Psi_G\big( w^T\Jo_G^cw \big)\\
							  & + \alpha\ \Psi_s( |\nabla_2\Dz|^2 ) d\Xy
\end{split}
\end{equation}

In order to estimate the solution, we apply Euler-Lagrange equation:
\begin{equation}
0 =  \bar{\Jo}_{11}\Dz + \bar{\Jo}_{12} - \alpha\ div\big([\Psi_s]' . \nabla_2 
\Dz \big)
\label{eq:eu_la_1}
\end{equation}
with the Neumann boundary condition $n^T[\Psi_s]'\nabla_2 \Dz = 0$. 
$\bar{\Jo}_{11}$, and 
$\bar{\Jo}_{12}$ refer to the two elements in the first row of joint 
light-field motion tensor defined as the following.
\begin{equation}
\bar{\Jo} = \operatornamewithlimits{\sum_{c=1}}^3 \Big( [\Psi_g^c]'  \Jo_g^c +  \gamma  [\Psi_G^c]'  \Jo_G^c \Big)
\end{equation}
where $[\Psi_{*}]'$ is the derivative of robustification function $\Psi(s)$. 
$[\Psi_{*}]' =\Psi_{*}'(s) = \frac{1}{2\sqrt{s+\epsilon_{*}}}$.

\subsection{Discretization}
We follow the discretization strategy from optical flow 
literature~\cite{Bruhn2006},~\cite{Zimmer2011} for spatial derivative. 
The directional derivative is computed as follows
\begin{equation}
L_{\Vs} = \frac{L(\Xy,\Vs_i) - L(\Xy,\Vs_0)}{|\Vs_i|}
\end{equation}

For each spatial discrete position $(i,j) \in \Omega$ we have
\begin{equation}
\begin{split}
&0 =[\bar{\Jo}_{11}]_{i,j}^{k}\Dz_{i,j}^{k+1} + [\bar{\Jo}_{12}]_{i,j}^{k} \\
   & - \alpha\sum\limits_{l\in x,y}\sum\limits_{(\tilde{i},\tilde{j})\in N_l} \frac{[\Psi_s']_{\tilde{i},\tilde{j}}^{k} + [\Psi_s']_{i,j}^{k} }{2} \bigg( \frac{\Dz_{ \tilde{i},\tilde{j}}^{k+1} - \Dz_{i,j}^{k+1} }{h_l^2} \bigg)
\end{split}
\label{eq:eu_la_dis_2}
\end{equation}
Here, $N_x=\{(i-1,j),(i+1,j)\}$, $N_y=\{(i,j-1),(i,j+1)\}$ and $h_x$,$h_y$ 
denote the unit distances for spatial derivative. The 
Eq.~\ref{eq:eu_la_dis_2} contains nonlinear term 
$\bar{\Jo}_{*}$,$\Psi_s'$. In order to solve it as a linear system of equation 
we apply lagged non-linearity method where $\bar{\Jo}_{*}$,$\Psi_s'$ are taken 
from the old time step ($k$) while computing $\Dz$ for the current time step 
($k+1$). By stacking Eq.~\ref{eq:eu_la_dis_2} for all pixels 
$(i,j)$ in spatial domain, we have a sparse $N\times N$-system of equations 
that could be solved using Gauss-Seidel iterative method. Here, $N=N_x \times 
N_y$ with $N_x$ and $N_y$ being spatial sampling resolution.

\subsection{Warping strategy}
\label{sec:warping}
As a condition for the linearization, we assume the displacement is small. This 
assumption will not always hold true, especially when we come to the directional 
position $\Vs_i$ that is far away from $\Vs_0$. In order to overcome this 
problem, we apply a coarse-to-fine warping technique~\cite{Bruhn2006}. The 
outline of our implementation is shown in  Algorithm.~\ref{alg:warping}.

The input parameters include the smoothness weight $\alpha$, the weight 
of gradient constancy data term $\gamma$, the down-sampling factor $\eta$, the 
standard deviation for Gaussian presmoothing $\sigma$ and the number of warping 
steps $l$.
We first initialize the set of down-sample light-field $L_i$ and resolution 
$R_i$ 
for each warping stage $i=1,...,l$. To downscale light-field, we first 
convoluted each sub-aperture image with Gaussian kernel and then sampled using 
bicubic interpolation. This is for avoiding alias during the down-sampling. We 
then begin with the coarsest level $l$ with the resolution of sub-aperture 
image 
down sample to $R_l = \eta^l(N_x,N_y)^T$. For each warping stage, the 
displacement from previous stage is up-scaled ($upscale$) to the current 
resolution. The light-field at the current level is warped to the central 
sub-aperture view ($warp$). The light-field motion tensors are then computed 
for intensity constancy ($lfmogray$) and gradient constancy ($lfmograd$) 
assumption. The disparity map $\Dz$ at current warping level is computed using 
the iterative solver ($itersol$).

\begin{figure}[t]
\begin{minipage}[b]{1.0\linewidth}
  \centering
  \includegraphics[width=\textwidth]{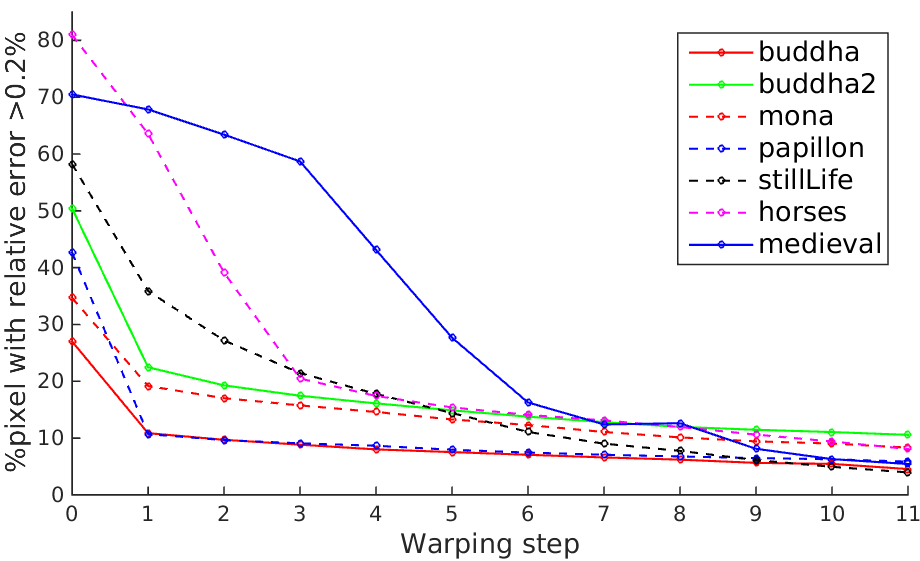}
\end{minipage}
\caption{Relative error after each warping level.}
\label{fig:warping}
\end{figure}

\begin{algorithm}
\caption{Warping strategy with coarse-to-fine}
\label{alg:warping}
\begin{algorithmic}[1]
\Require $\alpha$,$\gamma$,$\sigma$,$\eta$,$l$
\Procedure{DEPTH(L)}{}
\State initialize $R_i$,$L_i$
\State $\Dz = 0$
\For{($i=l; i \geq 0; i=i-1$)}
\State $\hat{\Dz} = upscale(\Dz,R_i)$
\State $\hat{L}_i = warp(L_i,\hat{\Dz})$
\State $\Jo_g = lfmogray(\hat{L}_i)$
\State $\Jo_G = lfmograd(\hat{L}_i)$
\State $\Dz = \hat{\Dz} + itersol(\Jo_g,\Jo_G,\alpha,\gamma)$
\EndFor
\Return $\Dz$
\EndProcedure
\end{algorithmic}
\end{algorithm}

Fig.~\ref{fig:warping} shows the changes in the percentage of erroneous 
pixels with relative error larger than 0.2\% after each warping stage. The 
results are computed for Synthetic light-field dataset~\cite{Wanner2013} with 
$l=11$ and $\eta=0.8$.  The graph shows great improvements on the disparity 
accuracy after each warping level.

\section{post processing}
\label{sec:post_processing}
Occlusion is one of the main problem in stereo matching. 
It happens when a patch of pixels exists in one view but vanishes in the others 
because of occluders.  
The problem is even more serious in the case of multi-stereo like light-field 
when 
the amount of vanish is varied from view to view. 
Our light-field motion tensors encode the information from all possible views 
and therefore contain least precise energy in occluded area. 
Since we compute the disparity map with respect to the central sub-aperture 
image, it makes sense to perform a post-processing to sharpen disparity field 
with respect to this reference view. 
For this purpose we propose a simple approach that employs the guided median 
filtering~\cite{He2013}.

\begin{figure}
\begin{minipage}[b]{\linewidth}
  \centering
  \begin{minipage}[b]{0.18\linewidth}
    \centering
    \includegraphics[width=\textwidth]{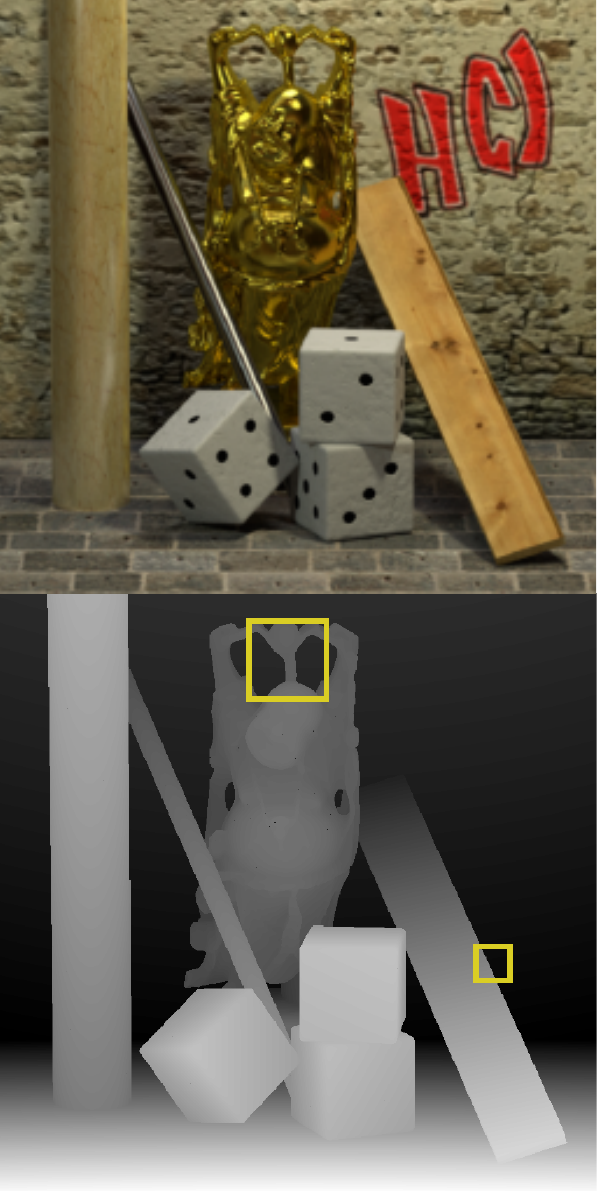}
    \centerline{(a)}\medskip
  \end{minipage}
  \hfill
  \begin{minipage}[b]{0.4\linewidth}
    \centering
    \includegraphics[width=0.9\textwidth]{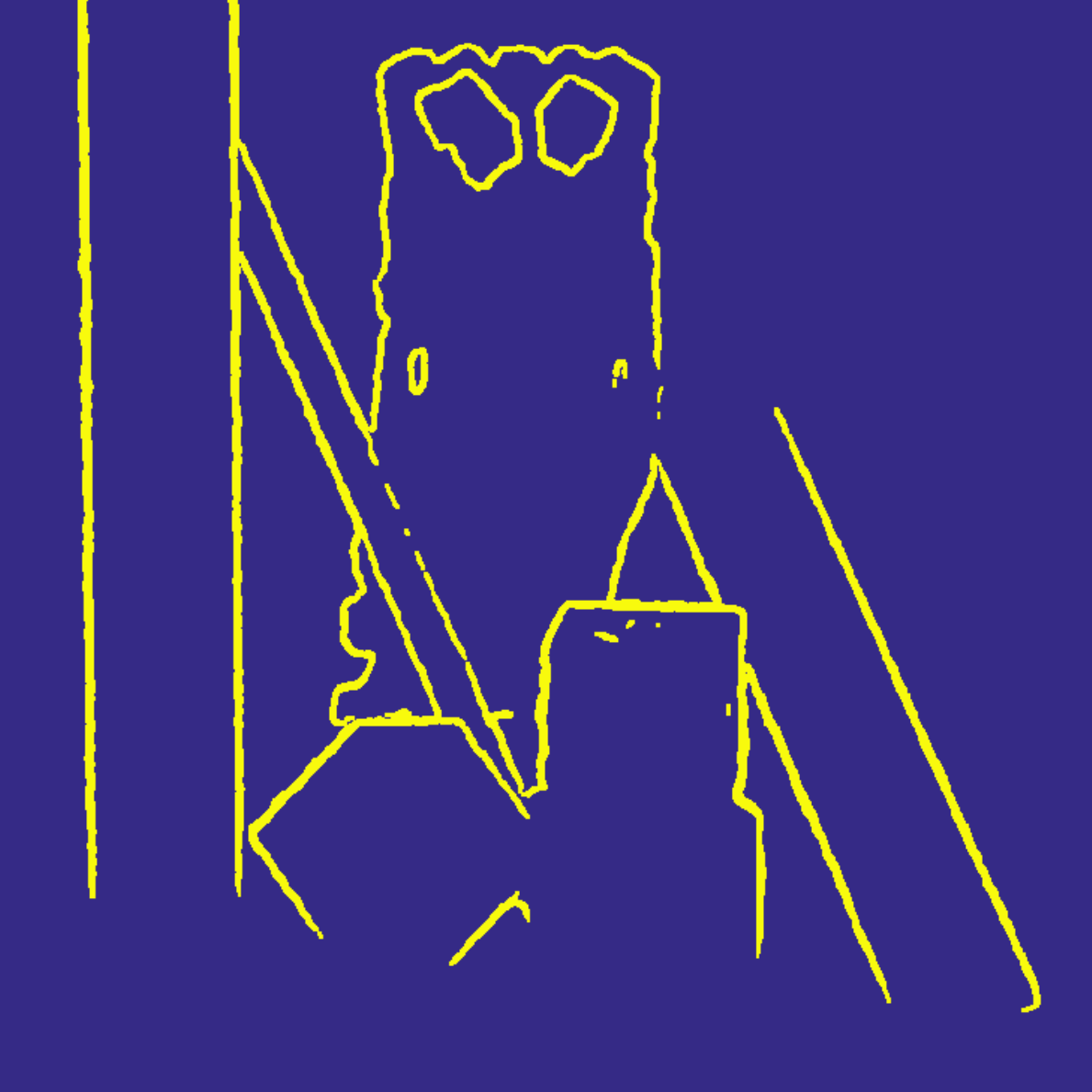}
    \centerline{(b)}\medskip
  \end{minipage}
  \hfill
  \begin{minipage}[b]{0.4\linewidth}
    \centering
    \includegraphics[width=1.1\textwidth]{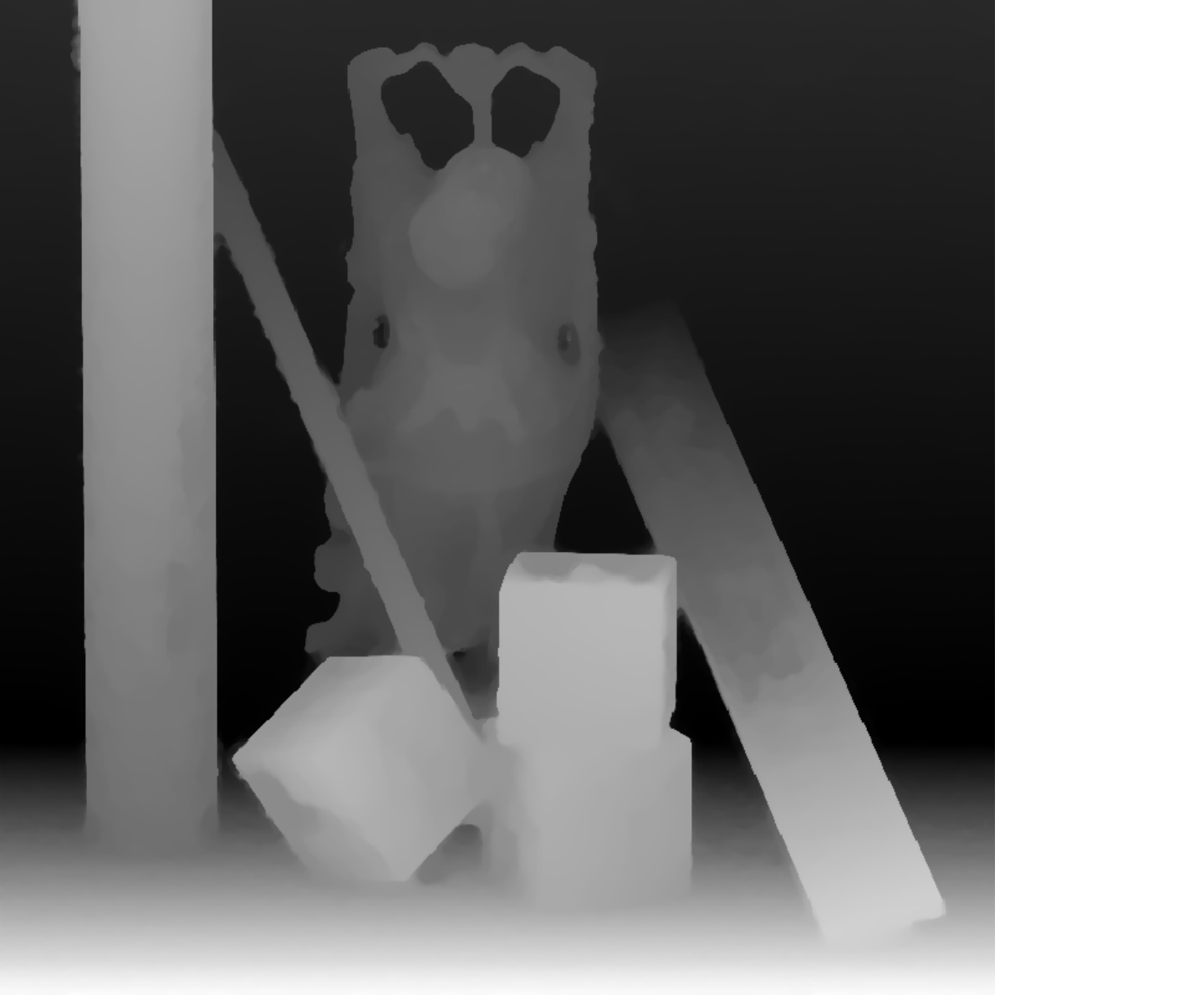}
    \centerline{(c)}\medskip
  \end{minipage}
\end{minipage}
\begin{minipage}[b]{\linewidth}
  \centering
  \begin{minipage}[b]{0.18\linewidth}
    \centering
    \includegraphics[width=\textwidth]{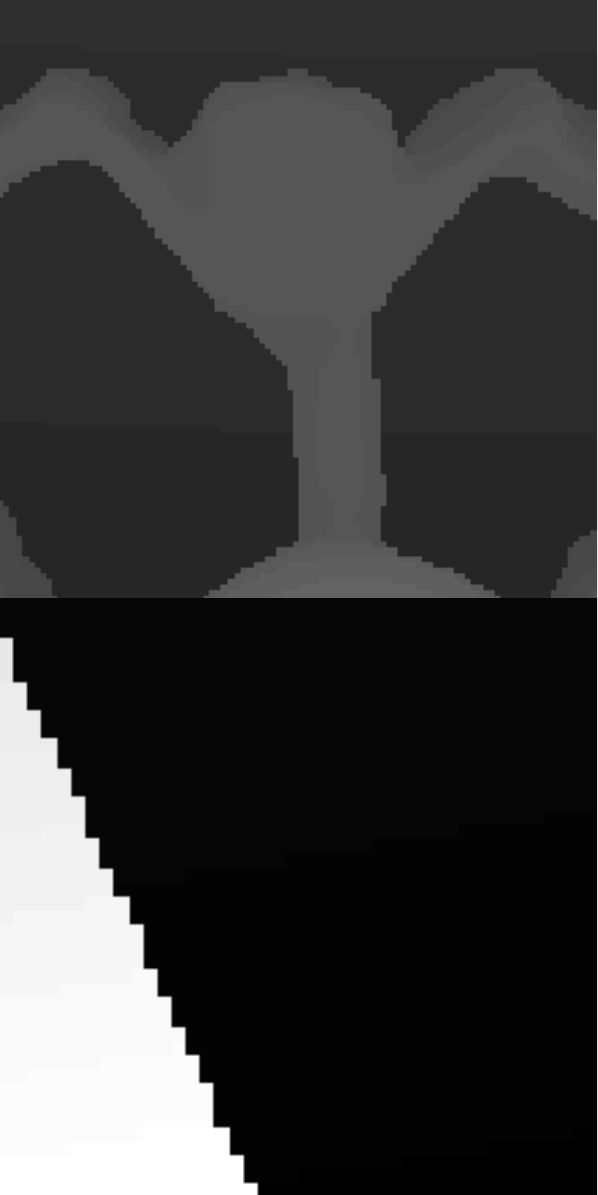}
    \centerline{(d)}\medskip
  \end{minipage}
  \hfill
  \begin{minipage}[b]{0.4\linewidth}
    \centering
    \includegraphics[width=0.9\textwidth]{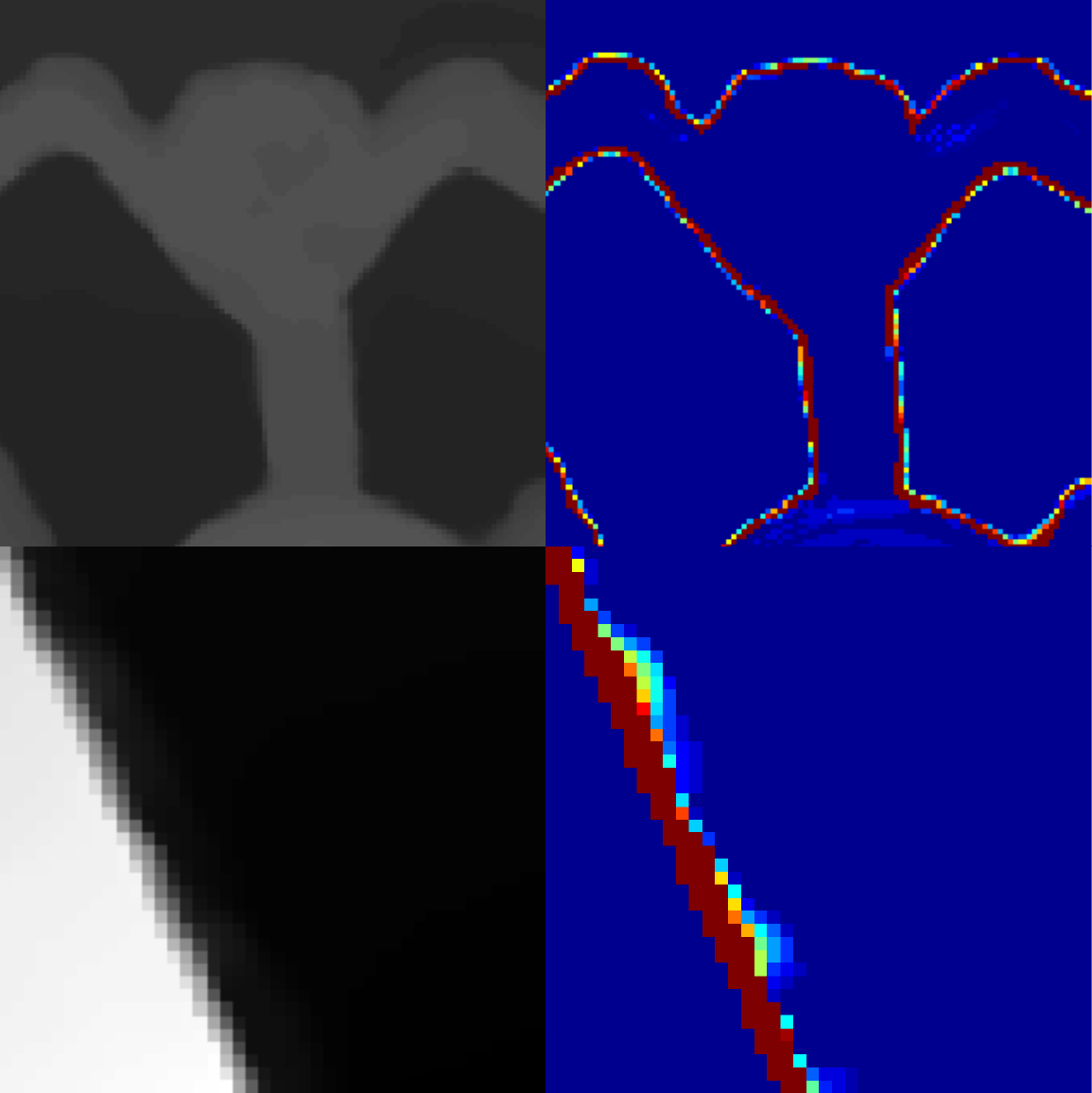}
    \centerline{(e)}\medskip
  \end{minipage}
  \hfill
  \begin{minipage}[b]{0.4\linewidth}
    \centering
    \includegraphics[width=1.1\textwidth]{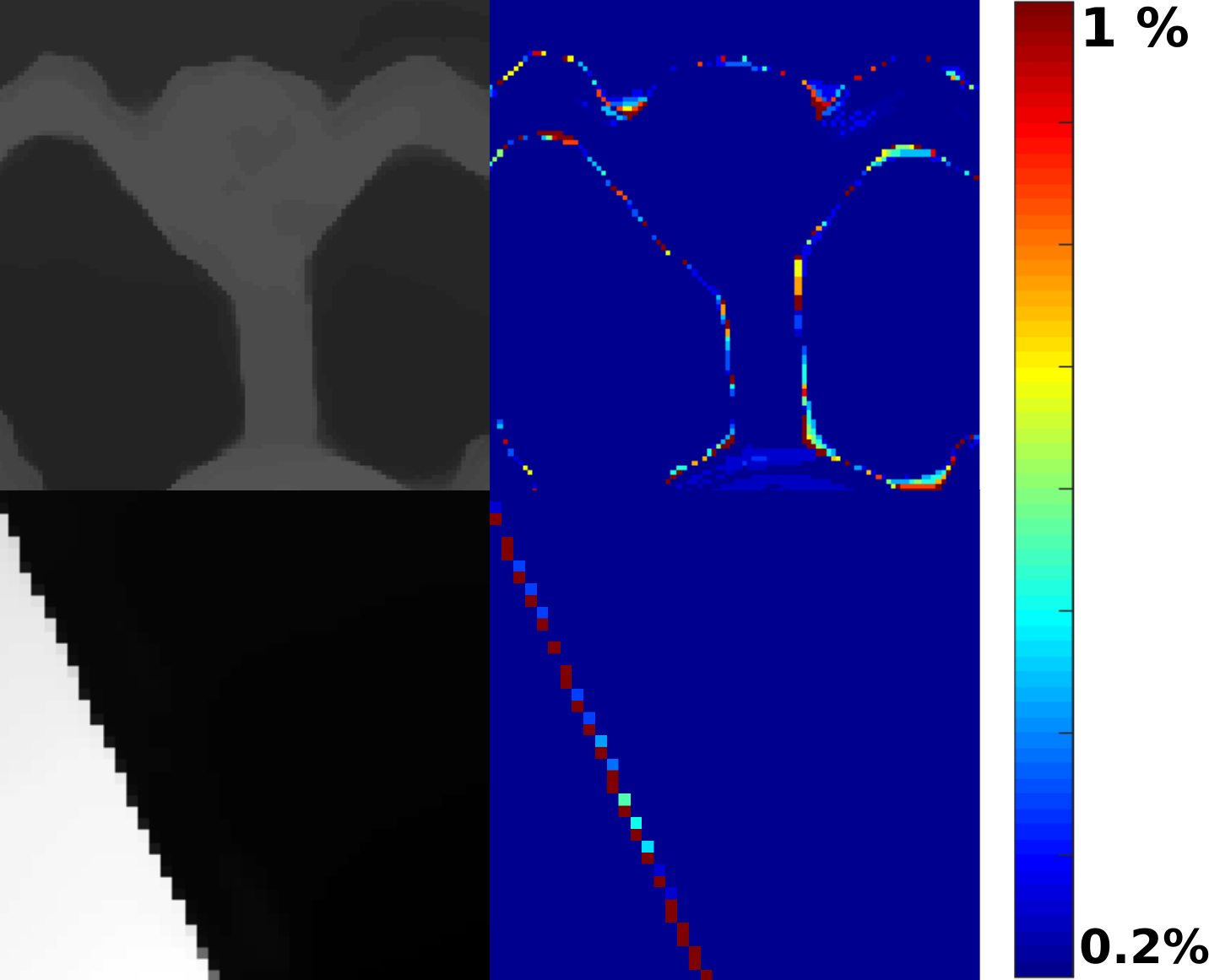}
    \centerline{(f)}\medskip
  \end{minipage}
\end{minipage}
\caption{Impact of proposed post-processing method on estimated disparity 
field. The result is computed for \textit{buddha2} scene from Synthetic 
dataset~\cite{Wanner2012}. 
(a) \textit{top}: the central sub-aperture image. \textit{bottom}: the ground 
truth with two interest areas.
(b) Estimated occlusion area.
(c) Computed disparity map.
(d) Zoom in ground truth for two interest areas.
(e),(f) estimated disparity and relative error before and after 
guided filtering respectively.}
\label{fig:post_processing}
\end{figure}

The procedure contains two steps. Firstly, we search for occluded areas where 
there is a high possibility of erroneous pixels. 
Secondly, we apply guided median filtering on the disparity map with 
respect to these occluded areas. We notice that the necessary condition for 
occlusion is depth discontinuity, and therefore we propose a simple occlusion 
detection based on computed displacement field. 

\begin{equation}
P_{occ} =  f(B_r \ast |\nabla_2\Dz|^2)
\end{equation}
where $f(\cdot)$ is a binary marking function that marks the response above 
some threshold and $B_r \ast $ denotes the convolution with a box kernel 
with size $r$ to expand suspected occlusion areas.
An example of marked occluded areas is shown in 
Fig.~\ref{fig:post_processing}(b). We then apply a median filter with 
a central 
sub-aperture image $Ic = L(\Xy,\Vs_0)$ as a guide. This allows us to have a 
sharper and more precise disparity discontinues as could be seen in 
Fig.~\ref{fig:post_processing}(e),(f).

\section{experimental results}
\label{sec:experiment}
The performance of the proposed variational framework was evaluated using both 
synthetic and real world datasets. The synthetic 4D Light-field 
dataset~\cite{Wanner2013} was used for quantitative comparisons with related 
work. All the real world dataset was captured using lenslet based light-field 
camera Lytro. We used both light-field data provided 
by EPFL~\cite{epfl2016} and data captured by our Lytro Illium camera.

We implemented our proposed framework on MATLAB. Source code of this 
implementation can be found at 
\url{https://github.com/hieuttcse/variational_plenoptic_disparity_estimation}  
. The computation ran on Intel i5 2.4Ghz CPU with 8GB RAM and required 4 
minutes for the \textit{Lytro} dataset and 9 minutes for the Synthetic dataset. 
We applied the Gauss-Seidel iteration solver with successive over-relaxation 
levels set to 1.88 which allowed the solution to converge faster, around 100 
iterations. 
The two parameters $\alpha,\gamma$ are adjusted for each light-field data. The 
other parameters were selected as $l=11$, $\eta = 0.8$, 
$\sigma =0.5$, $\epsilon_{*}=0.001^2$. 
 We notice that the most time consuming tasks are warping 
($warp(L_i,\hat{\Dz})$) and computing light-motion tensor ($lfmogray(\hat{L}_i), 
lfmograd(\hat{L}_i)$). These tasks could be dramatically speeded up on 
high-parallelism platform such 
as GPU or FPGA.

\begin{figure}[t]
\begin{minipage}[b]{1.0\linewidth}
\centering
\includegraphics[width=\textwidth]{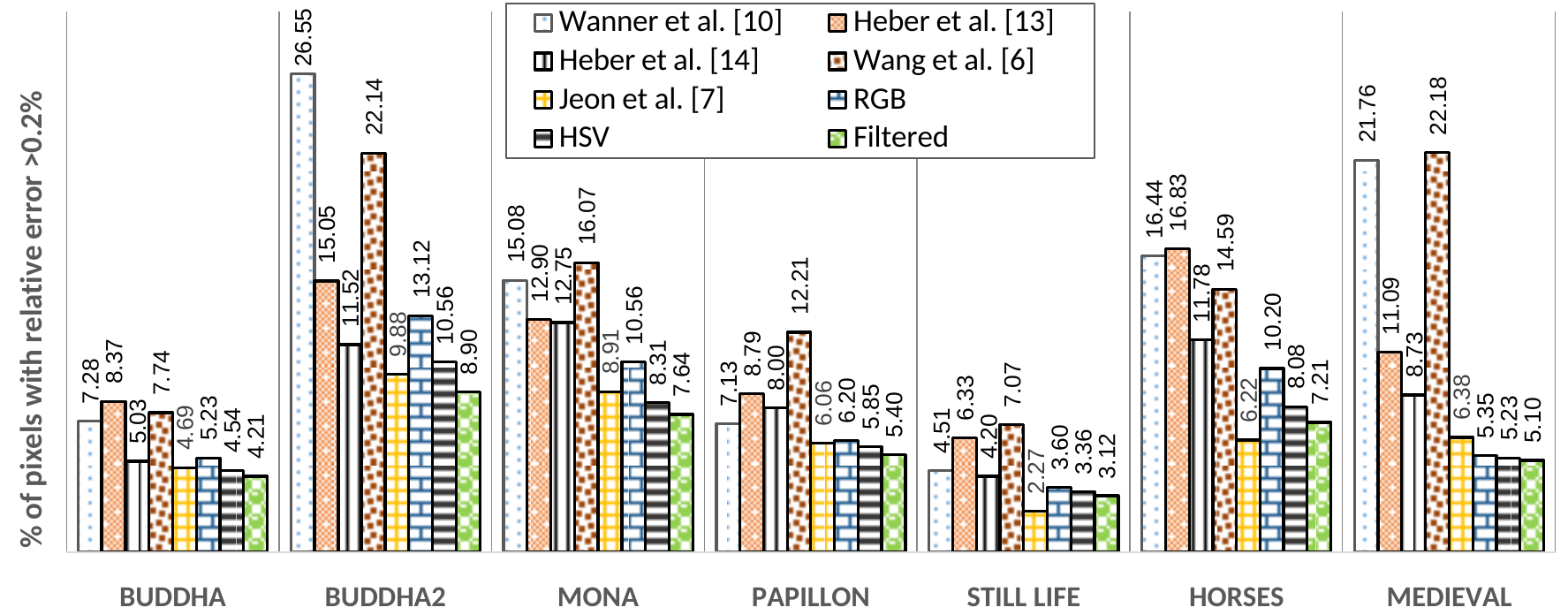}
\end{minipage}
\caption{Relative error comparison on synthetic dataset~\cite{Wanner2013}}
\label{fig:comparison}
\end{figure}

\subsection{Synthetic scene}
For quantitative evaluation, we compared our work with both continuous 
modelling approaches~\cite{Heber2013,Heber2014,Wanner2012} and 
discrete label-based approaches~\cite{Wang2015,Jaesik2015}. Since Wang 
et al.~\cite{Wang2015} did not report their results on synthetic dataset, we 
run their provided code for this comparison. For the others, we used their best 
results reported from related 
papers~\cite{Heber2013,Heber2014,Jaesik2015}.

\begin{figure*}
\begin{minipage}[b]{0.19\linewidth}
  \centering
  \begin{minipage}[b]{\linewidth}
    \centering
    \includegraphics[width=\textwidth]{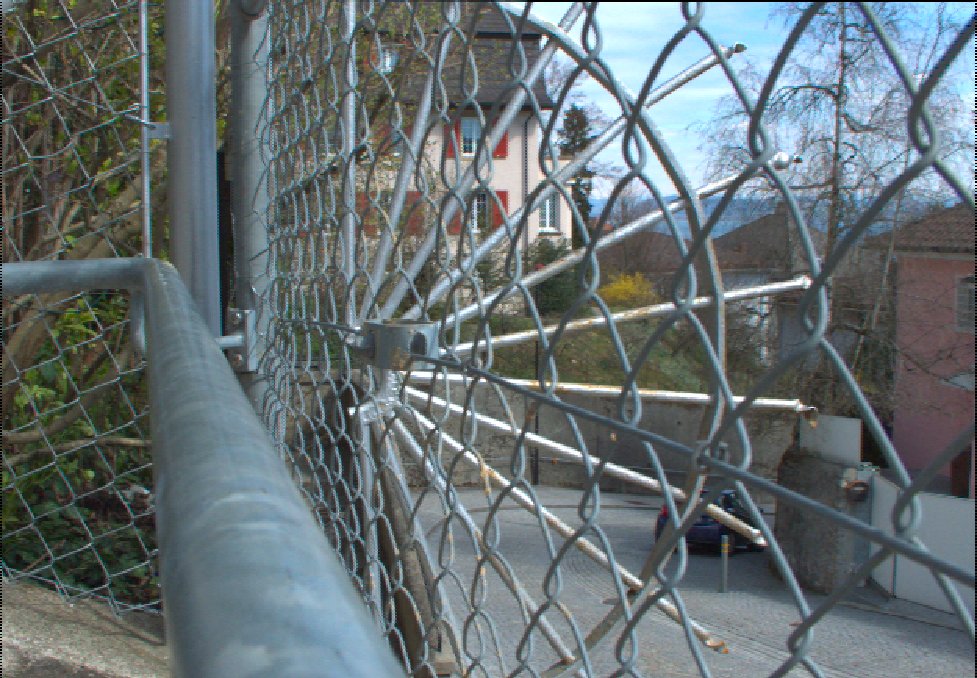}
  \end{minipage}
  \begin{minipage}[b]{\linewidth}
    \centering
    \includegraphics[width=\textwidth]{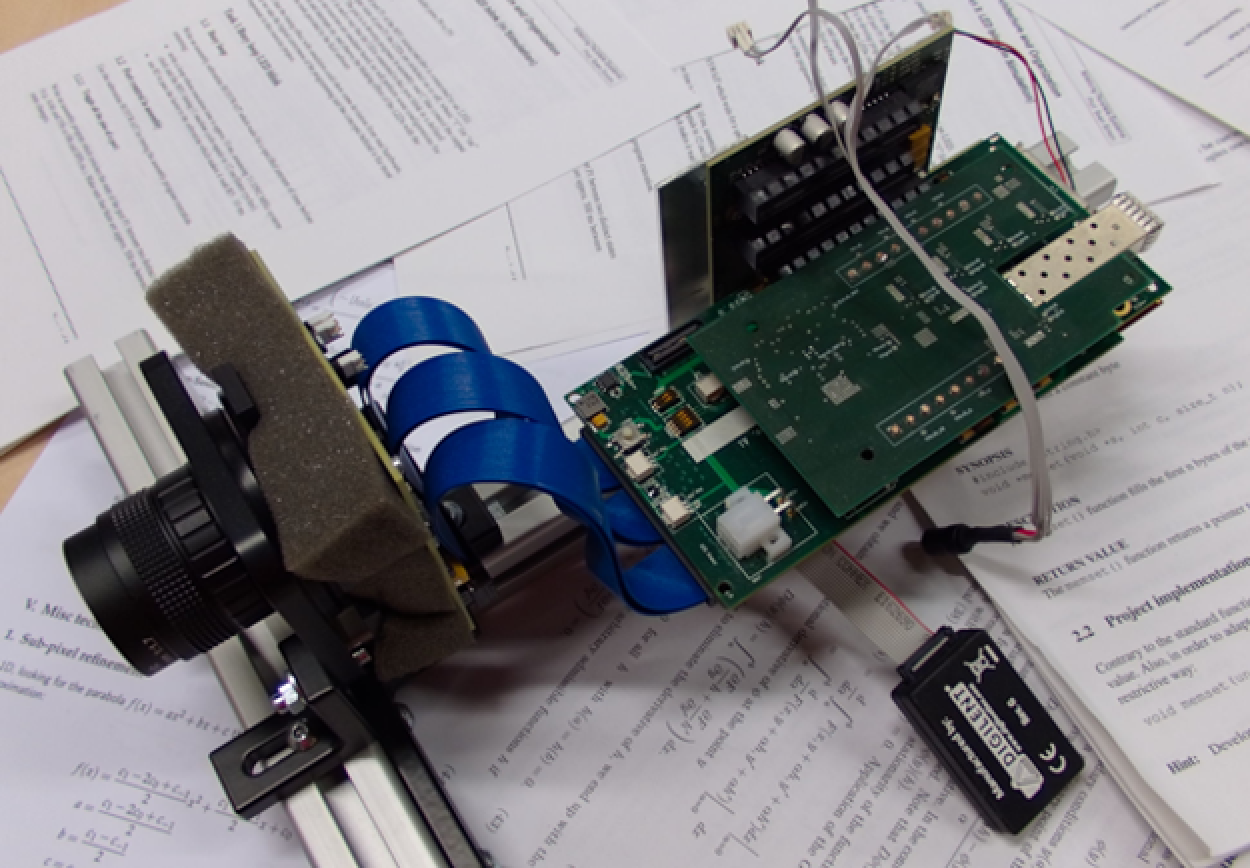}
  \end{minipage}
  \centerline{Center view}\medskip
\end{minipage}
\hfill
\begin{minipage}[b]{0.19\linewidth}
  \centering
  \begin{minipage}[b]{\linewidth}
    \centering
    \includegraphics[width=\textwidth]{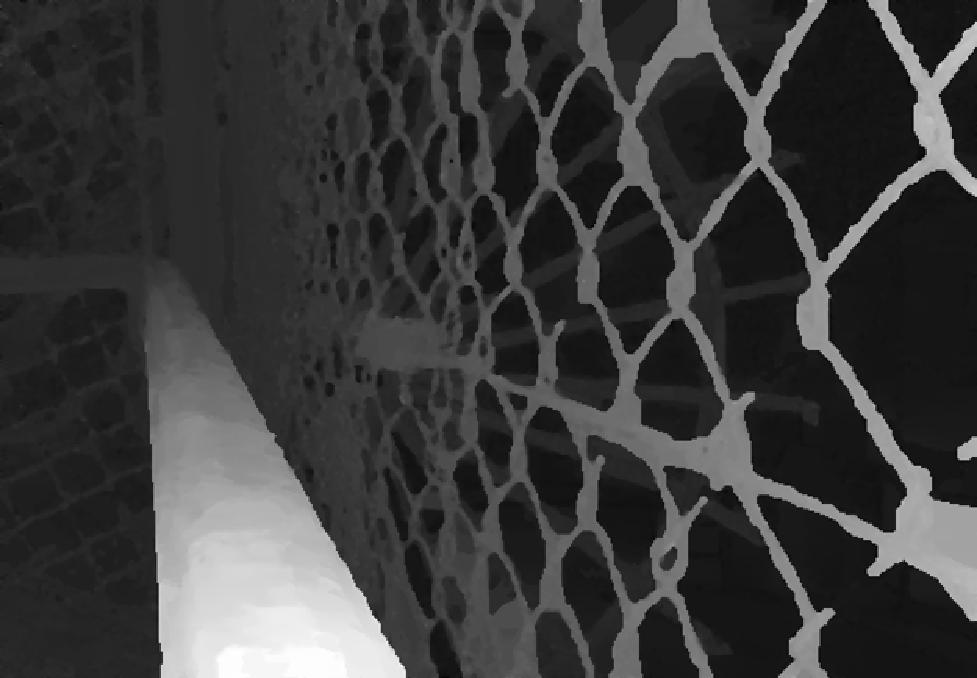}
  \end{minipage}
  \begin{minipage}[b]{\linewidth}
    \centering
    \includegraphics[width=\textwidth]{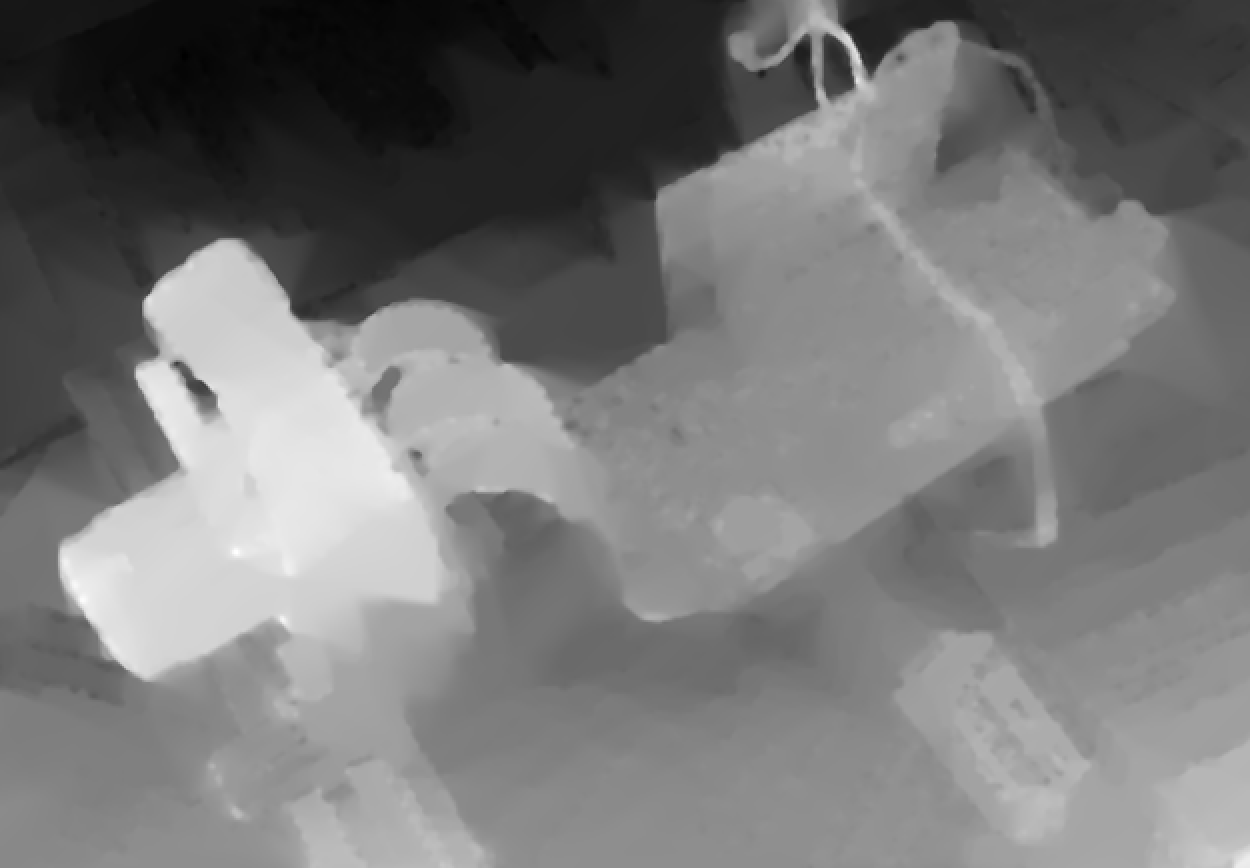}
  \end{minipage}
  \centerline{Lytro software}\medskip
\end{minipage}
\hfill
\begin{minipage}[b]{0.19\linewidth}
  \centering
  \begin{minipage}[b]{\linewidth}
    \centering
    \includegraphics[width=\textwidth]{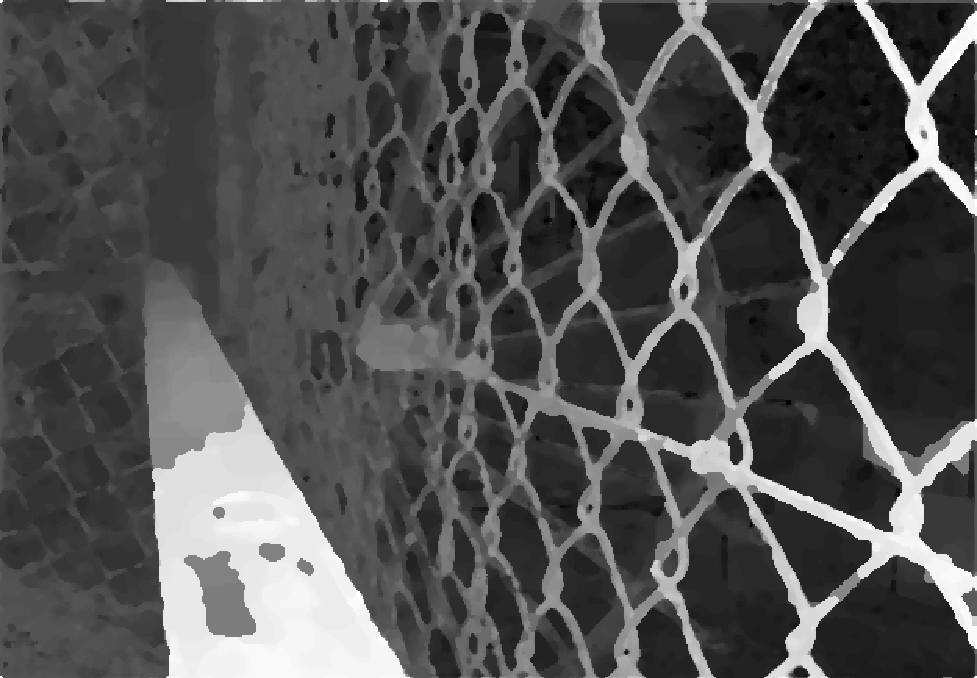}
  \end{minipage}
  \begin{minipage}[b]{\linewidth}
    \centering
    \includegraphics[width=\textwidth]{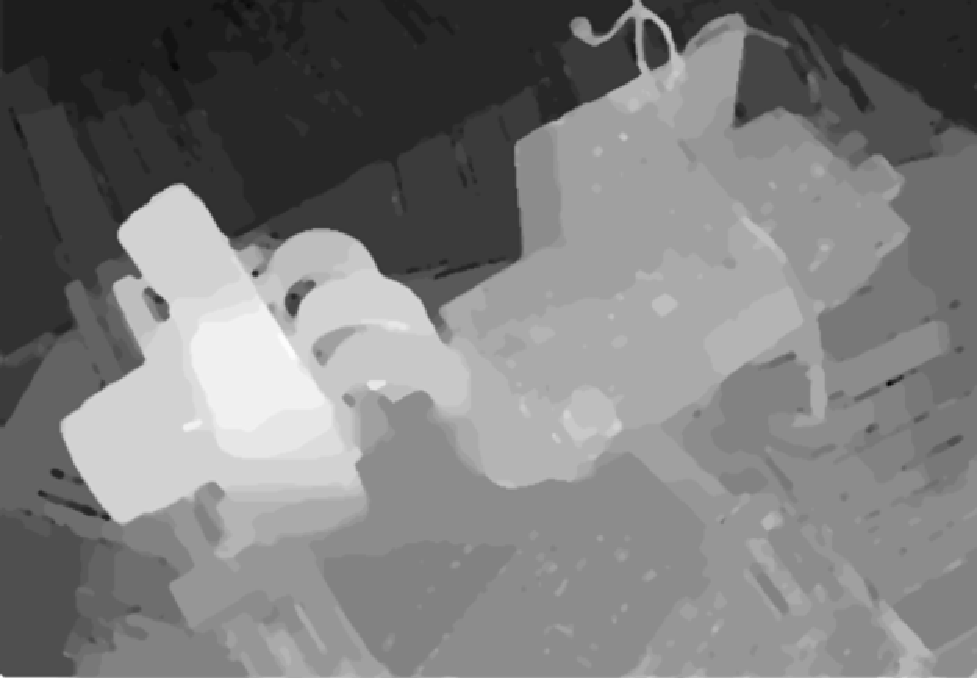}
  \end{minipage}
  \centerline{Wang et al.~\cite{Wang2015}}\medskip
\end{minipage}
\hfill
\begin{minipage}[b]{0.19\linewidth}
  \centering
  \begin{minipage}[b]{\linewidth}
    \centering
    \includegraphics[width=\textwidth]{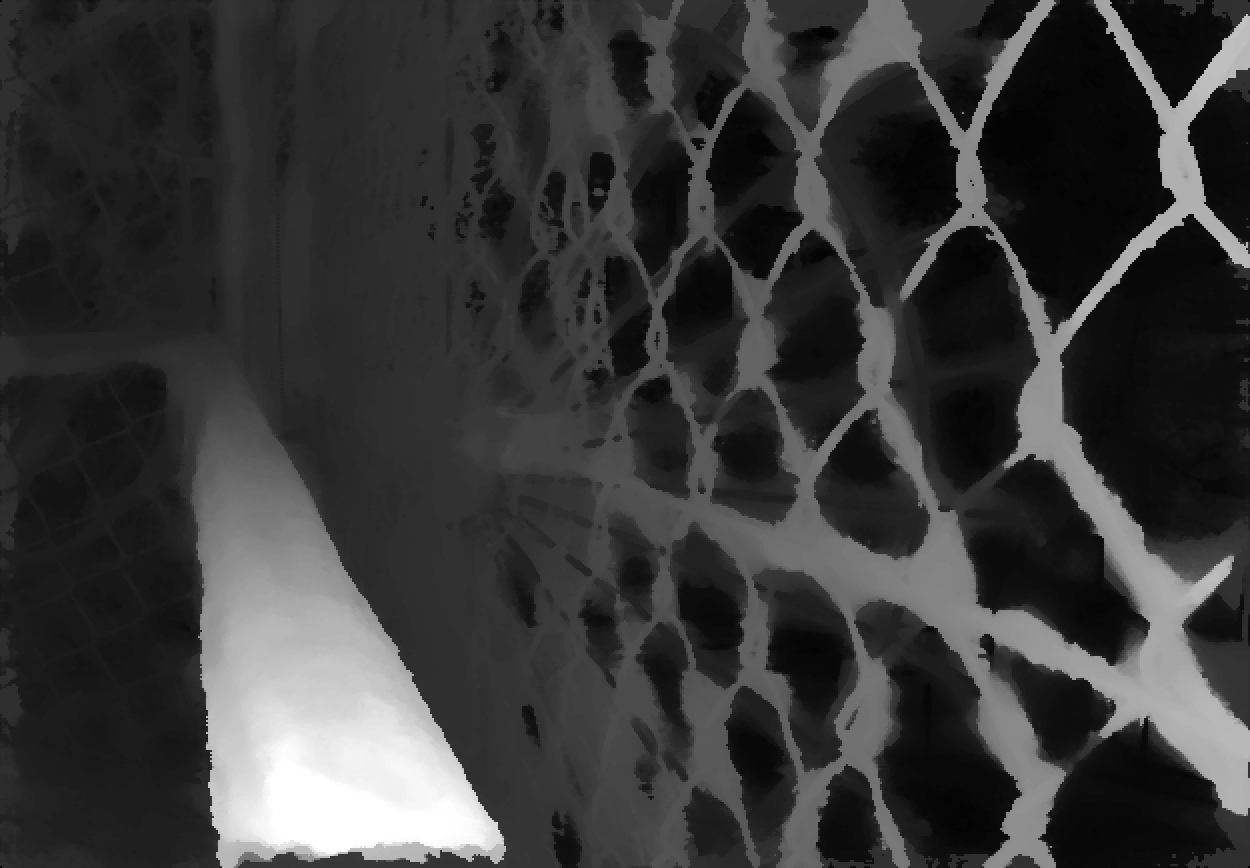}
  \end{minipage}
  \begin{minipage}[b]{\linewidth}
    \centering
    \includegraphics[width=\textwidth]{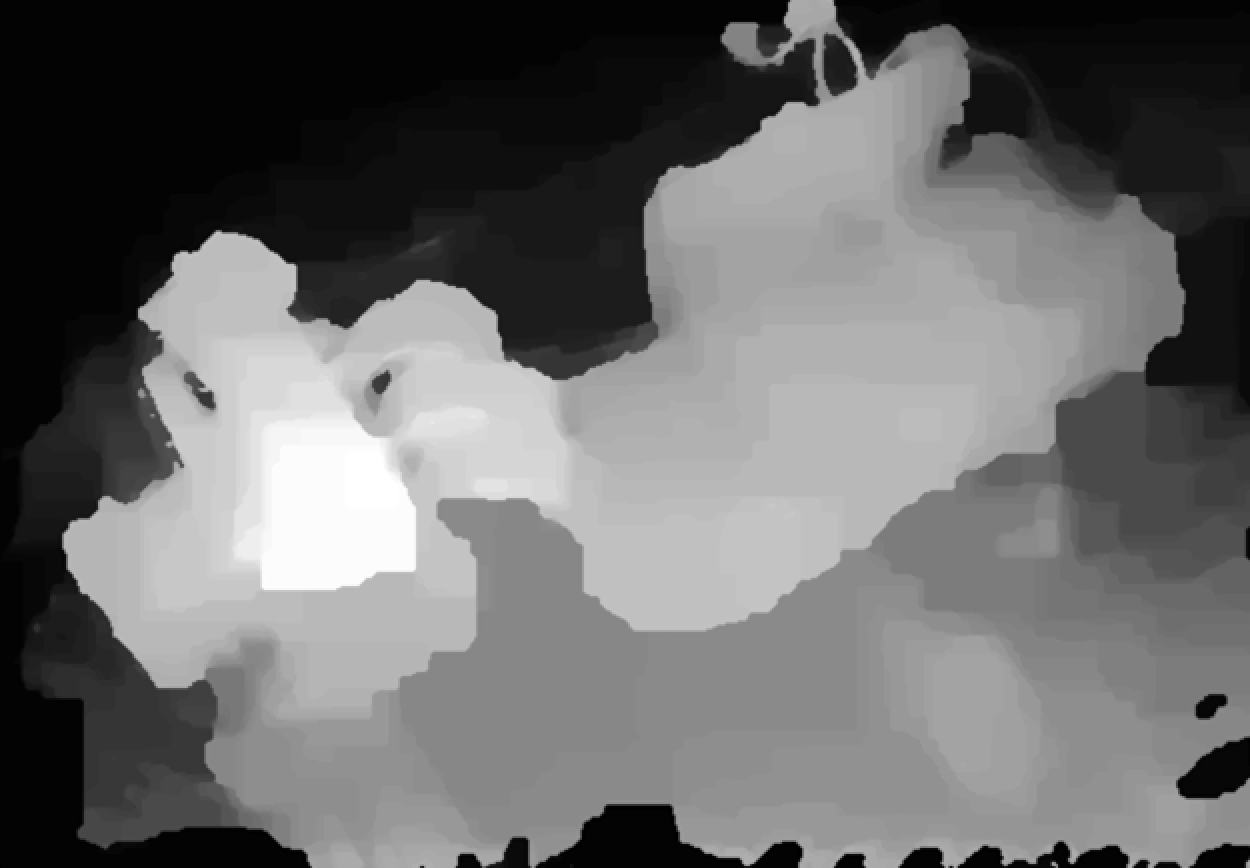}
  \end{minipage}
  \centerline{Jeon et al.~\cite{Jaesik2015}}\medskip
\end{minipage}
\hfill
\begin{minipage}[b]{0.19\linewidth}
  \centering
  \begin{minipage}[b]{\linewidth}
    \centering
    \includegraphics[width=\textwidth]{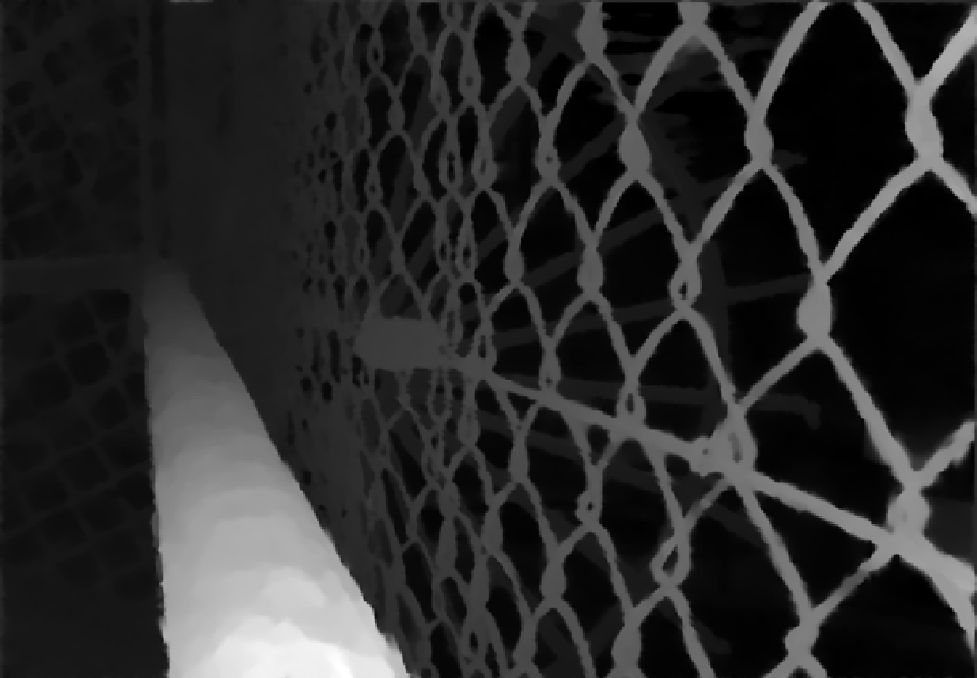}
  \end{minipage}
  \begin{minipage}[b]{\linewidth}
    \centering
    \includegraphics[width=\textwidth]{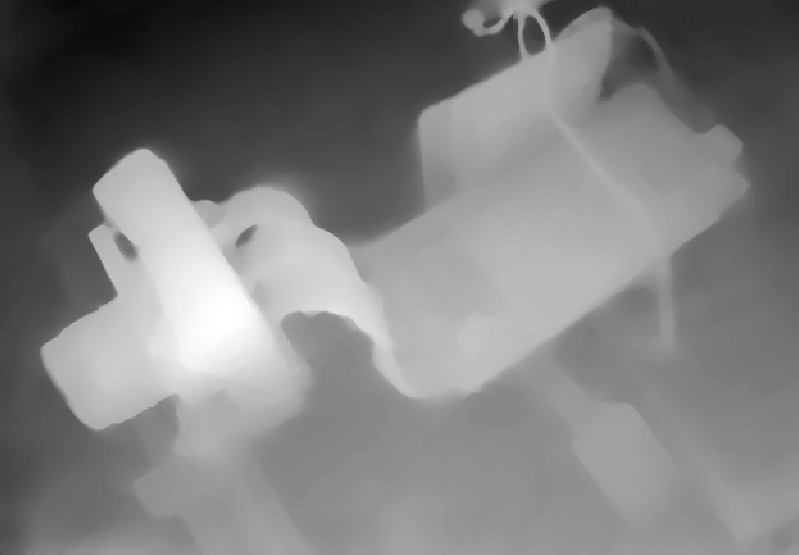}
  \end{minipage}
  \centerline{Proposed approach}\medskip
\end{minipage}
\caption{Comparison on real-world dataset. \textit{top} scene from EPFL 
dataset~\cite{epfl2016}. \textit{bottom} scene from our captured 
light-field. }
\label{fig:real_compare}
\end{figure*}

The relative error is adopted as a common metric for comparison. 
For each synthetic light-field data, we computed the percentage of pixels with 
a relative depth error of more than 0.2\%. As mentioned by 
Heber~\cite{Heber2013}, this is the smallest meaningful accuracy level since 
the depth discretization of the provided ground truth is too low. We reported 
the result for both RGB and HSV color space and the post-processing result of 
the best solution.

Fig.~\ref{fig:comparison} shows the relative error results of the previous 
works and ours. It can be seen from the bar chart that our proposed framework 
outperforms previous continuous modelling 
approaches~\cite{Heber2013,Heber2014,Wanner2012} in term of 
accuracy. Compared with the RGB color space, the computation on the HSV color 
space 
provides more precision due to the separate robustification of the data term. 
The guided median filtering further improves the accuracy by sharpening the 
disparity discontinues. We notice that the approach of Wang et 
al.~\cite{Wang2015} is very sensitive to estimated occlusion results, and tends 
to have a wrong disparity estimation at strong texture areas. Their approach 
therefore is less accurate when compared to the others. 

Compared with the work of Jeon et al.~\cite{Jaesik2015}, our approach 
provides better results for the five out of seven synthetic light-field 
data. The two less accurate results (\textit{still life},\textit{horses}) show 
some insights about the limitation of our approach that will be discussed at the 
end of this section. However, advantages of our approach are as follows. 
Firstly, our approach does not require the previous knowledge 
of disparity range as well as spatial disparity unit for each disparity label 
due to the continuous formation. Without this knowledge, discrete label-based 
approaches~\cite{Wang2015},\cite{Jaesik2015} need to increase the number of 
labels and reduce the spatial unit in order to sufficiently cover the disparity 
range. It consequently increases the computation time. In addition, a number of 
well-turned parameters for each data, around 9 parameters, are critical in 
their 
approach~\cite{Jaesik2015} in order to guarantee a smooth disparity field 
without outliers. Our 
framework, in contrary, requires mainly adjustment on only two parameters 
$\gamma$ and $\alpha$. 

\subsection{Real scene}
In this section, we provided some qualitative results on real world dataset 
obtained by using \textit{Lytro} camera. 
The images are captured using Illium version which provides a spatial 
resolution of around $434 \times 625$ and a directional resolution of $15 
\times 15$. Due to the vignetting impact of microlens, the effective 
directional 
resolution is limited to only 193 views (85.7\%). Fig.~\ref{fig:real_common} 
presents the results of our proposed approach for different scenes. 
Along with the central sub-aperture image, there is also a rendered 3D view 
as well as a color-coded disparity map. Fig.~\ref{fig:real_compare} shows the 
disparity map for two different scenes computed respectively by the Lytro 
software, the 
methods of Wang et al.\cite{Wang2015}, Jeon et al.\cite{Jaesik2015} and ours. 
The results for~\cite{Wang2015,Jaesik2015} are computed using their 
provided 
codes. It can be seen from the figure that our proposed approach provides 
solutions with more details and fewer outliers. 

\subsection{Limitation and future work}
The accuracy of our approach is limited by the error at depth discontinues 
where occlusion takes place. While guided median filtering does improve the 
accuracy at these areas, it could also introduce ``halo'' 
artefact~\cite{He2013} when the texture borders in guided image are weak. One 
solution for this problem is taking into account the occlusion 
border~\cite{Wang2015} to neglect the artefact. Another solution is to employ 
more sophisticated smoothness terms that could better preserve the depth 
discontinues such as image- and flow-driven regularizer~\cite{Zimmer2011} or 
Total Generalized Variation (TGV)~\cite{Ranftl2012}. These tasks are listed in 
our plan for future work. 

\section{conclusion}
\label{sec:conclusion}
In this paper we proposed a variational computation framework for the disparity 
estimation problem targeting very narrow-baseline multi-stereo data from 
plenoptic images.
We introduced a light-field motion tensor that allows different constancy 
assumptions to be applied.
We embedded coarse-to-file warping strategy to our framework in order to 
overcome the 
problem of large displacement and proposed an effective post-processing 
technique for further enhancing the accuracy at occluded areas. The 
experimental 
results show our competitive performance on both challenging synthetic and real 
world light-field dataset.

\begin{small}
\bibliographystyle{IEEEbib}
\bibliography{lightfield}
\end{small}

\end{document}